\documentclass[a4, 10pt, conference, final]{ieeeconf}
\usepackage{amsmath}
\usepackage[utf8]{inputenc}
\usepackage{siunitx}
\usepackage[hidelinks]{hyperref}
\usepackage{graphicx}
\usepackage{subcaption}
\usepackage{tikz}
\usepackage[capitalise]{cleveref}
\usepackage[nolist]{acronym}
\usepackage{svg}
\usepackage{mathtools}
\usepackage{pdfpages}
\usepackage{amsmath,amsfonts,amssymb}
\usepackage[normalem]{ulem}
\usepackage{nicefrac}
\usepackage{makecell}
\usepackage{relsize}
\usepackage{multirow}
\usepackage{leftidx}
\usepackage{tabularx}
\usepackage[percent]{overpic}
\usepackage[noadjust]{cite}
\usepackage{dblfloatfix} 
\usepackage{booktabs}
\usepackage[font=small,labelfont=bf]{caption}

\usetikzlibrary{positioning}

\newcommand{\comment}[1]{}

\IEEEoverridecommandlockouts

\definecolor{clr_detectionMeasurement}{rgb}{1,0,1}
\definecolor{clr_detectionEstimatedMean}{rgb}{0.4,0.4,0.4}

\definecolor{clr_darkgreen}{rgb}{0.0, 0.6, 0.33}

\newcommand\hl[1]{%
  \bgroup
  \hskip0pt\color{green!70!black}%
  #1%
  \egroup
}

\newcommand\hlx[1]{%
  \bgroup
  \hskip0pt\color{cyan!90!black}%
  #1%
  \egroup
}

\begin{document}
\title{%
A Multi-Vehicle Dataset with Camera, LiDAR, and Radar Sensors and Scanned 3D Models for Custom Auto-Annotation using RTK-GNSS
}
\author{Philipp Berthold\textsuperscript{*}, Bianca Forkel\textsuperscript{*}, and Mirko Maehlisch\thanks{The authors are with the Institute for Autonomous Driving of the University of the Bundeswehr Munich, Neubiberg, Germany. Contact author email: \{philipp.berthold, tas\}@unibw.de.
Dataset and supplementary data available: 
\url{https://github.com/UniBwTAS/7V-Scanario}.
}}

\maketitle

\begingroup\renewcommand\thefootnote{*}
\footnotetext{Philipp Berthold and Bianca Forkel contributed equally to this work.}
\endgroup

\begin{abstract}
Datasets are a crucial element in the development of perception algorithms.
They relate sensor measurement data to annotated reference information
and allow for the deduction of sensor and object characteristics.
In autonomous driving, the reference data commonly consist of semantic image segmentation, point-wise associations, or bounding box annotations.
The dataset proposed in this work, however, aims to dig deeper into the evaluation of measurement principles and provides scanned 3D models of all vehicles together with a pose and continuous kinematics reference obtained by RTK-GNSS.
Combined, the state of the complete dynamic surrounding of the sensor vehicle is known for any point in time.
Subsequent reference formats can be easily computed in user-defined granularity.
This dataset involves single-object and multi-object recordings with seven target vehicles.
In particular, measurement effects such as occlusion, as well as reflections, can be evaluated, as the normals of the shape of the target vehicles are known.
We describe the dataset, discuss the technical background of its development, and briefly present exemplary evaluations.
\end{abstract}

\IEEEpeerreviewmaketitle

\section{Introduction}
\label{cha:Introduction}
The development of perception algorithms for autonomous driving strongly benefits from datasets, since they support the modeling and learning of measurement principles and object appearances in sensor data.
Furthermore, they allow for the evaluation of different algorithms and parameter choices.

Reference data (``ground truth'' information) about other objects in the environment is often provided in form of 2D or 3D bounding boxes (e.g., \cite{Argoverse2}), which lack information about the object's boundaries.
More detailed reference data is supplied by pixel-wise semantically (sem.) annotated images for camera sensors (e.g., \cite{Geiger2012CVPR,nuscenes2019,Sun_2020_CVPR,Cordts2016Cityscapes}) or by point-wise object annotation for LiDAR (e.g., \cite{behley2019iccv,fong2021panoptic}) or radar point clouds (e.g., \cite{Schumann2021radarscenes}).
Ideally, the annotations contain (unique) instance (inst.) IDs to be able to distinguish between different object instances or even track one object over time.
However, the association of a point to a certain instance ID is sometimes ambiguous as the point might be caused by a reflection, i.\,e., results from the mutual interaction between multiple objects.
Only in \cite{Menze2015}, CAD models of the specific vehicle types are fitted to the sensor data to give additional information about the objects' shape and structure.
This fitting is challenging as there is no positional reference available for the third-party vehicles; therefore, only the sensor data can be used to determine the poses of other objects.
Furthermore, the used publicly available, generic CAD models do not necessarily conform to the true shape of the observed objects.
Thus, the fitted CAD models cannot achieve particularly high accuracy.
The CAD models are also only available in the depth images used as ground truth information in a stereo benchmark.
Overall, as \cref{tab:related_work} shows, existing datasets are missing detailed 3D models of the exact observed vehicles.

Besides, annotated reference data is usually only provided every few seconds.
In addition, from sensor data annotations, no precise information about the object's kinematics (like velocity and rotational speed) can be derived.
Thus, \cite{Schlichenmaier2022} gains reference data from GNSS-based inertial navigation systems that are installed on the objects. This way, the dataset can serve as ground truth for radar-based object tracking algorithms.
However, as shown in \cref{tab:related_work}, there is no kinematic reference data of the target vehicles in datasets with camera images or LiDAR point clouds.

\begin{table*}
    \caption{
    Comparison of our dataset to popular or related autonomous driving datasets with respect to dynamic objects.
    }
    \label{tab:related_work}
    \centering
    \scriptsize
    \setlength{\tabcolsep}{5.8pt}
    \begin{tabular}{l|cccc}
        \toprule
        & Sensors & Annotations & Kinematics & 3D Models \\
        \midrule
        (Semantic) KITTI \cite{Geiger2012CVPR} & camera, LiDAR & \makecell{2D/3D boxes, sem./inst. image segm., \\ sem./inst. LiDAR segm. \cite{behley2019iccv}} & sensor vehicle & \makecell{simple CAD models \\ in depth images \cite{Menze2015}} \\
        (Panoptic) nuScenes/nuImages \cite{nuscenes2019} & camera, LiDAR, (radar) & \makecell{2D/3D boxes, sem./inst. image segm., \\ sem. LiDAR segm. \cite{fong2021panoptic}} & sensor vehicle & no \\
        Cityscapes (3D) \cite{Cordts2016Cityscapes} & camera & 3D boxes \cite{gählert2020cityscapes3ddatasetbenchmark}, sem./inst. image segm & sensor vehicle & no \\
        Waymo \cite{Sun_2020_CVPR} & camera, LiDAR & 2D/3D boxes, sem./inst. image seg., sem. LiDAR seg. & no & no \\
        Argoverse 2 \cite{Argoverse2} & camera, LiDAR & 3D boxes & no & no \\
        Oxford (Radar) RobotCar \cite{RobotCarDatasetIJRR} & (camera, LiDAR, radar \cite{RadarRobotCarDatasetICRA2020}) & INS & sensor vehicle \cite{RCDRTKArXiv} & no \\
        GOOSE \cite{Mortimer2024goose} & camera, LiDAR, (radar) & sem./inst. image segm., sem./inst. LiDAR segm. & sensor vehicle & no \\
        RadarScenes \cite{Schumann2021radarscenes} & (camera), radar & sem./inst. radar point labels for moving objects & sensor vehicle & no \\
        \cite{Schlichenmaier2022} & radar & IMU+RTK-GNSS solution & target vehicles & no \\
        \textbf{7V-Scanario (Ours)} & camera, LiDAR, radar & user-defined auto-annotation & all vehicles & 3D scans \\
        \bottomrule
    \end{tabular}
\end{table*}

\textbf{Contributions:} 
This work aims to contribute a dataset containing camera, LiDAR, and radar measurements and providing continuous kinematic reference data together with precise 3D structure and texture information for every object.
Every vehicle in this dataset is equipped with an RTK-GNSS/INS unit computing precise positional and kinematic information.
Additionally, a 3D scan has been performed for every object involved.
This scan resolves the structure of the object and also contains the object's texture (color information).
Every detail of the vehicle, including the installation of the GNSS/INS unit, is thus known.
\begin{figure}
    \centering
    \includegraphics[width=1\linewidth,trim={5cm 1.5cm 0cm 3.5cm},clip]{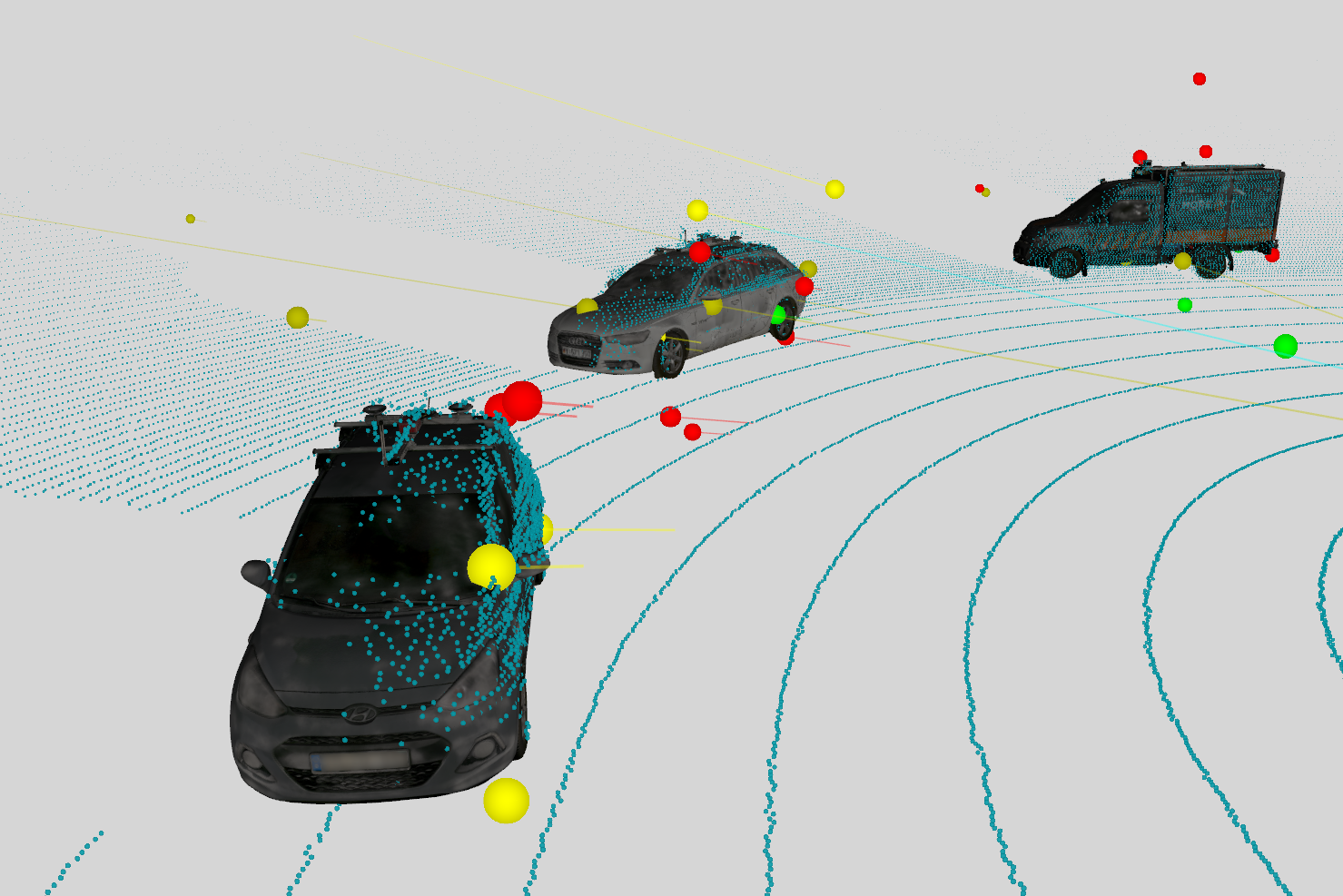} 
    \caption{The scanned 3D models of three vehicles and the measurement data (blue spheres: LiDAR point cloud, yellow/red/green spheres: radar detections with Doppler measurements denoted by lines) rendered live in \emph{rviz} (ROS middleware). \vspace{-.5em}}
    \label{fig:e87_rviz}
\end{figure}
Combined, full 3D information of the vehicles around the sensor vehicle can be obtained for any given point in time and put into relation with the measurement data (\Cref{fig:e87_rviz}).

By respective processing, reference data like range images or pixel/point annotations can be generated.
\begin{figure}
    \centering
    \includegraphics[width=1\linewidth,trim={5cm 0cm 7cm 4cm},clip]{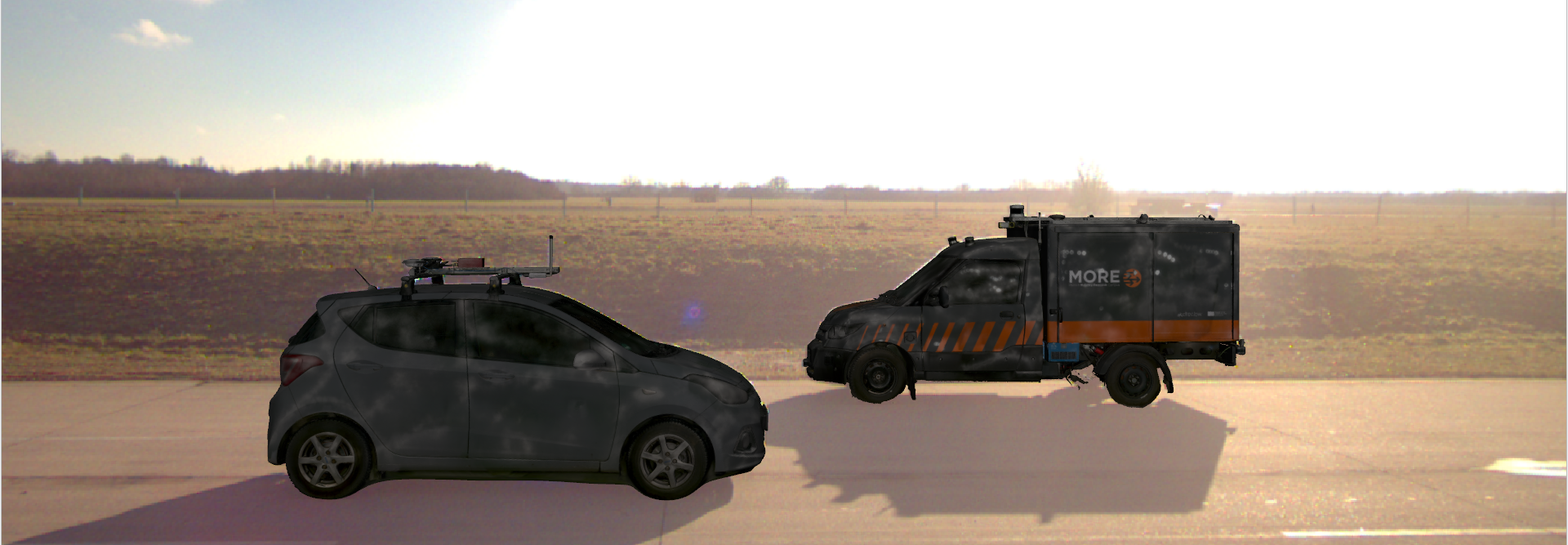} 
    \caption{The projection of the scanned 3D models into a camera image showing the original vehicles. The projection error is barely recognizable. By modifying the texture of the model, image data augmentation can be easily performed.}
    \label{fig:i10_projektion}
\end{figure}
\begin{figure}[b]
    \centering
    \begin{subfigure}{.49\linewidth}
        \centering
        \includegraphics[width=\linewidth]{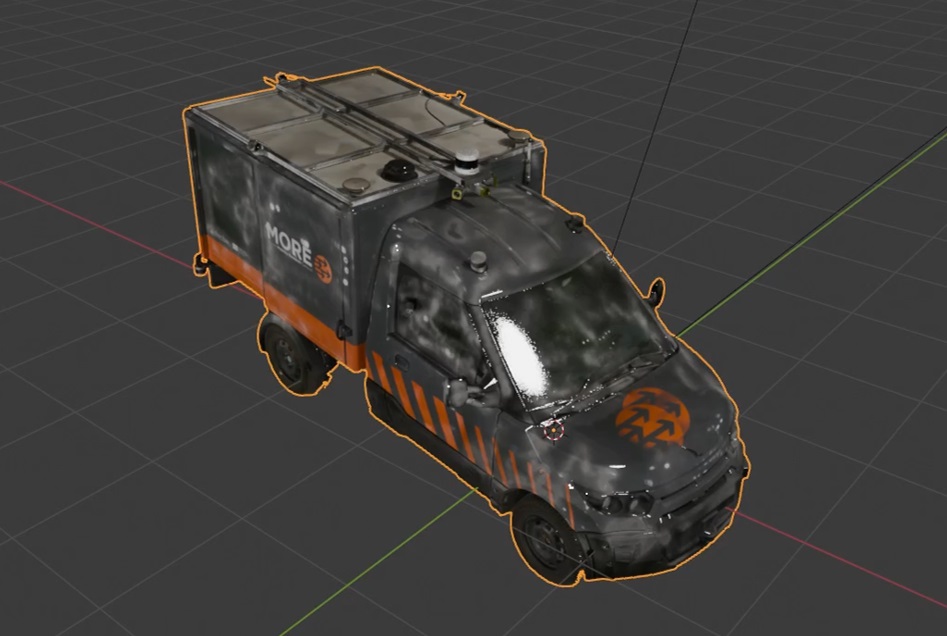}
    \end{subfigure}%
    \hfill
    \begin{subfigure}{.49\linewidth}
        \centering
        \includegraphics[width=\linewidth]{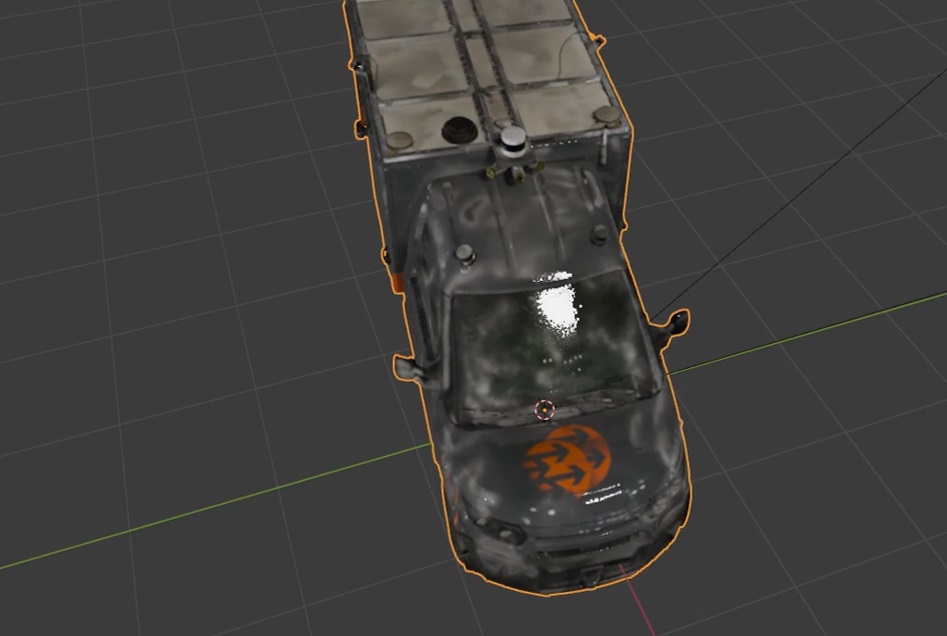}
    \end{subfigure}%
    \caption{The illumination of the scanned 3D models with artificial sources of light can also be used to extend image data augmentation.}
    \label{fig:sts_licht}   
\end{figure}
Since the dataset contains very detailed 3D scans, the reference data can be generated in any granularity. For example, a pixel or point can be annotated not only as a car, but also more specifically as a wheel or window. This is of great importance for taking physical measurement principles into account.
For example, in radar sensors, wheels generate micro-Doppler measurements.
Rather than manually labeling all sensor data, it is sufficient to decompose the scanned 3D object into the desired granularity once, thereby enabling automatic annotations for all sensor data.
Moreover, advanced data augmentation techniques like changing the texture of the objects (\Cref{fig:i10_projektion}) or altering or adding sources of light (\Cref{fig:sts_licht}) can be employed to extensively enhance the training data concerning camera images.
The proposed dataset concept can also serve to ``split'' the Sim-to-Real Gap by capturing a clinical environment with real measurement data and comprehensive ground truth information and providing the real 3D models for simulation.

Our dataset comprises seven heterogeneous vehicles and two sets of scenarios. The first set of scenarios focuses on fundamental measurement principles: each vehicle is measured separately from various distances and viewing angles on a large, open, flat paved surface. Both the sensor and target vehicle are moving to obtain Doppler measurements for the radar sensors. The second set consists of multi-object scenarios that vary from dynamic and narrow trajectories, causing frequent occlusions, to realistic crossing and highway scenarios on an empty test track.

\textbf{Structure:} 
This paper is structured as follows: \Cref{cha:features} denotes the principal features and contents of this dataset. \Cref{cha:setup} gives insight into the technical background of the setup. \Cref{cha:examples} outlines some exemplary evaluations based on this dataset, and \Cref{cha:conclusion} finally summarizes this paper.

\section{Dataset Features}
\label{cha:features}

\subsection{Sensors}
\begin{figure}[b]
    \centering
    \includegraphics[width=1\linewidth]{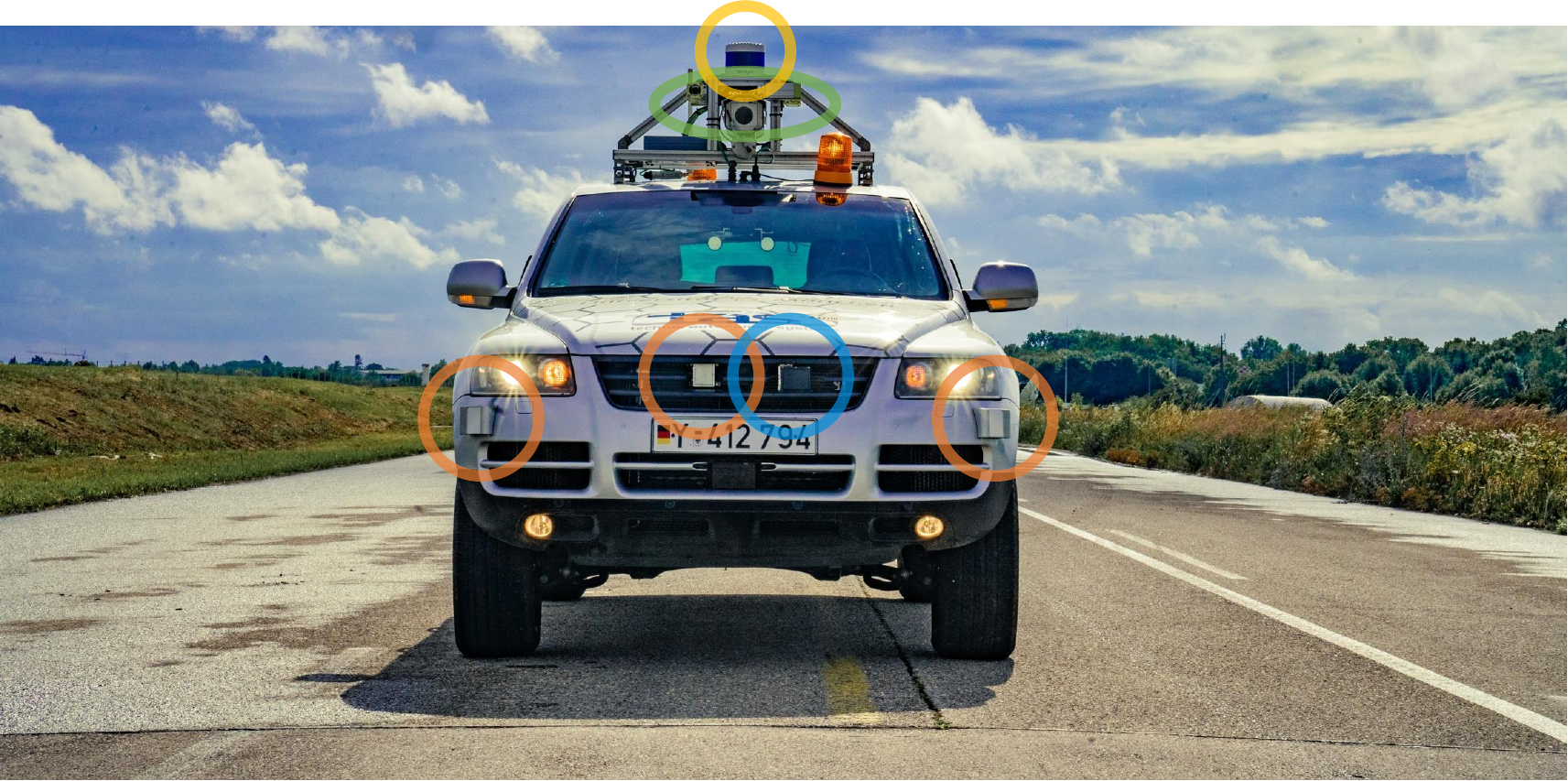}
    \caption{The front of the sensor vehicle \emph{MuCAR-3} (\emph{VW Touareg}). The yellow circle denotes the LiDAR sensor, the green circles the cameras, the orange circles the medium range radar sensors, and the blue circle denotes the far range radar sensor.}
    \label{fig:sensor_setup_front}
\end{figure}
\begin{figure*}
    \centering
    \includegraphics[width=1\linewidth]{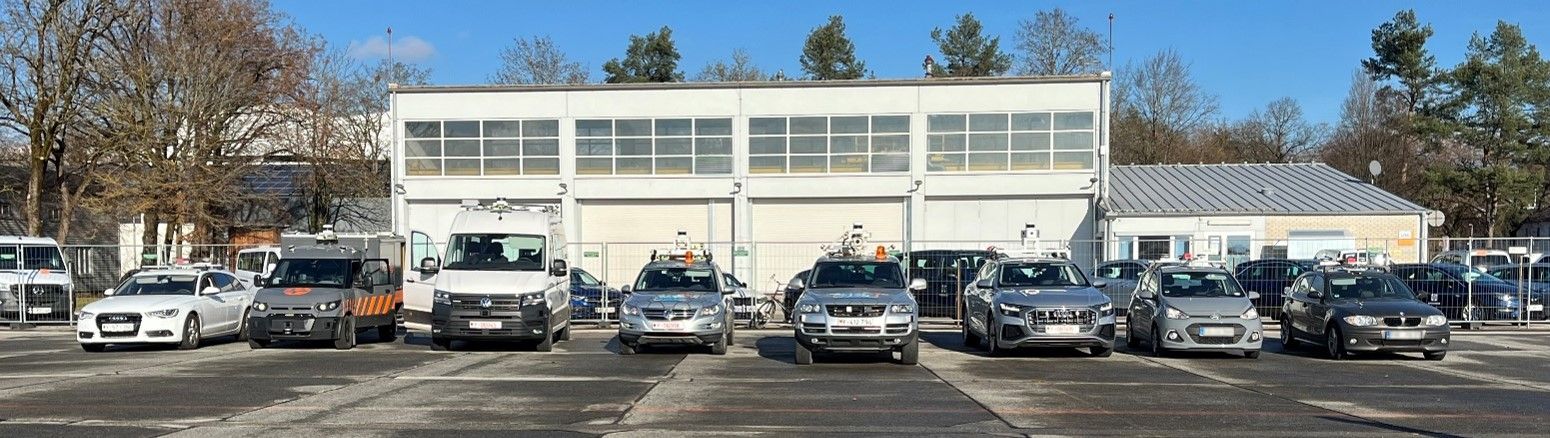}
    \caption{All vehicles of this dataset (targets and sensor vehicle). From left to right: \emph{Audi A6} station wagon, \emph{StreetScooter Work} delivery vehicle, \emph{Volkswagen e-Crafter} transporter, \emph{VW Tiguan} SUV with various sensors on the roof, \emph{VW Touareg} (utilized as sensor vehicle), \emph{Audi Q8} SUV with various sensors on the roof, \emph{Hyundai i10} compact car, \emph{BMW 1 series} compact car.}
    \label{fig:vehicles}
\end{figure*}
The dataset was recorded using the research vehicle \emph{MuCAR-3} (cf. \cite{Mortimer2024goose}), which is a modified \emph{Volkswagen Touareg}.
\Cref{fig:sensor_setup_front} depicts its front and some of the mounted sensors.
In this dataset, we provide continuous and complete raw data (without any filtering or postprocessing) of the following sensors:
\begin{itemize}
    \item 1x LiDAR sensor \emph{Velodyne VLS-128}: mounted at the center of the vehicle, providing \SI{360}{\degree} point clouds at \SI{10}{\hertz}.
    \item 4x RGB cameras \emph{Basler acA2440-20gc}: mounted in vicinity of the LiDAR sensor, spaced in \SI{90}{\degree} intervals and providing images at \SI{10}{\hertz}.
    \item 5x Radar sensor \emph{Smartmicro UMRR-96 Type 153}: \SI{79}{\giga\hertz} medium range radar, mounted on the four corners and at the radiator grill and providing 4D detections at roughly \SI{18}{\hertz}.
    \item 1x Radar sensor \emph{Smartmicro UMRR-32 Type 132}: \SI{77}{\giga\hertz} far range radar, mounted at the radiator grill and providing 4D detections at roughly \SI{18}{\hertz}.
    \item 1x RTK-GNSS/INS unit \emph{OxTS RT3000v3}: using correction data from a local reference station (less than \SI{1}{\kilo\meter} distance in all scenarios) and providing inertial measurement data and absolute positional information (with an accuracy of about \SI{1}{\centi\meter}, see \cref{{sec:accuracy}}) at \SI{100}{\hertz}. Similar RTK-GNSS/INS units (either \emph{OxTS RT3000v3} or \emph{OxTS RT3003}) are utilized on the target vehicles to obtain their ground truth poses.
    \item Series vehicle data of the \emph{VW Touareg}: providing wheel speeds, steering angle, and gyro data at \SI{50}{\hertz}.
    This data is used to deduct a jump-free odometry (by estimating the thermal drift using a Kalman filter).
\end{itemize}
The documentation of the GOOSE \cite{Mortimer2024goose} (German Outdoor and Offroad Dataset) dataset, which uses the same sensor setup, gives more detailed information on the sensor vehicle and the sensors.
The sensors have been calibrated extrinsically by the probabilistic joint multi-sensor optimization algorithm described in \cite{Forkel2025calibration} to yield preferably accurate mounting poses (detailed per-sensor evaluations in \cite{Forkel2025calibration}). 
In addition, hardware timestamping, hardware triggering, and software-based bus latency compensation relate all sensor data to a common clock.
As the RTK-GNSS/INS units provide reference data at a high frequency, ground truth data for any arbitrary sensor measurement time can be computed by interpolation. The data of all units is shared using long-range WiFi and dumped to local integrated storages.

\subsection{Vehicles}
The dataset comprises seven heterogeneous target vehicles (see \Cref{fig:vehicles}), starting with compact cars up to large delivery vans.
Each vehicle has been equipped with an RTK-GNSS/INS unit so that its exact position is known at all times. In addition, a detailed 3D scan is available for all vehicles. Furthermore, for each vehicle, a dedicated single-object recording was performed.

\subsubsection{Hyundai i10}
This car represents the subcompact car category. It has been equipped with a roof structure that carries two GNSS multi-band multi-frequency antennas and the INS itself. Care has been taken for preferably low impact on the sensor measurements, although it cannot be completely avoided. However, as is valid for all vehicles, the scan covers this structure so that the measurements match the 3D model. The coordinate system of the 3D model corresponds to the local coordinate system of the INS (identical reference point and rotation).

\subsubsection{BMW 1 Series}
This car represents compact cars. It has been equipped with an identically constructed roof structure.

\subsubsection{Audi A6 station wagon}
This car represents station wagons. It has been equipped with a roof construction identical to that of the two previous models.

\subsubsection{VW Tiguan}
This vehicle is an SUV-based prototype vehicle (\emph{MuCAR-4}). It carries different sensors, and thus possesses a rather complex roof structure. The INS is permanently mounted inside the vehicle.

\subsubsection{Audi Q8}
This vehicle is also an SUV-based prototype vehicle (\emph{MuCAR-5}), and has a roof structure with various sensors. The INS is installed inside the vehicle.

\subsubsection{StreetScooter Work}
This is a prototype vehicle for automated logistics and is based on a delivery vehicle. It carries some sensors along its roof and at its corners and an INS unit inside. A peculiarity is its plastic transport box that is moderately permeable to radar waves. As additionally multiple reflections may occur inside this box, clutter detections can be measured at positions far beyond the transport box (in radial view of the sensor).

\subsubsection{VW e-Crafter}
This vehicle is a transporter with a long wheelbase. It was also converted into a prototype vehicle and comes with some sensors on the roof, the front and on the corners, and an INS installed in the vehicle.

\subsection{Scenarios}
\setcounter{figure}{7}
\begin{figure*}[b]
    \centering
    \includegraphics[trim=0 130 0 100,clip,width=1\linewidth]{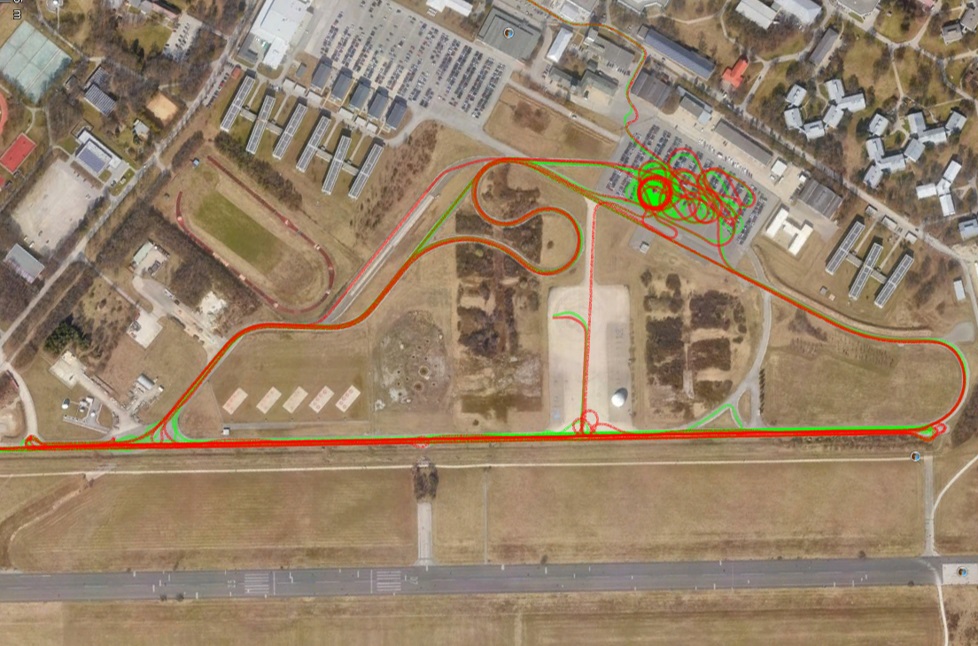}
    \caption{Aerial view of the path of the sensor vehicle (red) and exemplarily the \emph{VW Tiguan} (green) covering all multi-object scenarios. The test track contains crossings, circular courses, country roads, a long multi-lane passage (at the bottom, formerly used as an airport taxiway), and open spaces.}
    \label{fig:multiobjekt_pfad}
\end{figure*}
\subsubsection{Single-object}
The purpose of this scenario set is the observation of the single target vehicles from various distances and angles. To this end, in every recording, the sensor vehicle circles around one target vehicle, changing the radius over time. During this process, the target vehicle is slowly moving forward to provoke Doppler measurements and to obtain diverse ground and background appearances.
\setcounter{figure}{5}
\begin{figure}
    \centering
    \includegraphics[width=1\linewidth]{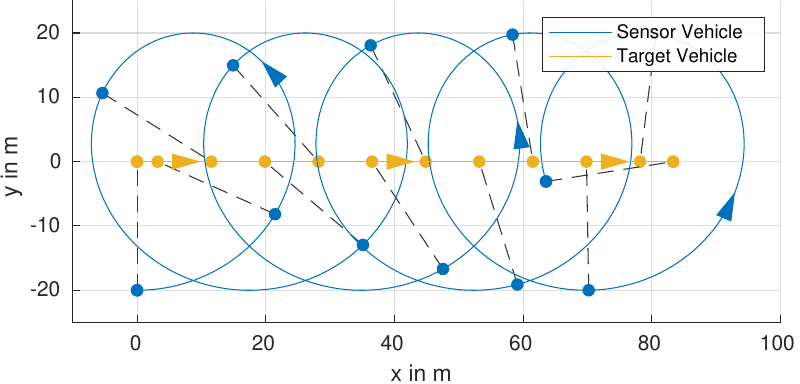}
    \caption{Schematic trajectories of sensor and target vehicle of a single-object recording: the sensor vehicle moves along circular paths around the target vehicle, which is slowly moving forward.}
    \label{fig:trajektorie}
\end{figure}
\Cref{fig:trajektorie} illustrates both trajectories.
\begin{figure}
    \centering
    \includegraphics[angle=270,width=1\linewidth]{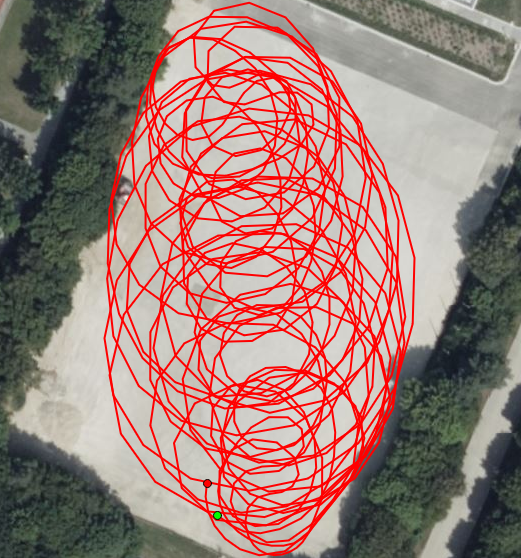}
    \caption{Driven path of the sensor vehicle on an asphalted free space during a single-object recording (here: \emph{VW e-Crafter}). This free space has an area of \SI{110}{\meter} by \SI{70}{\meter}. The driven circles have radii from \SI{7}{\meter} up to \SI{35}{\meter}.\vspace{-1em}}
    \label{fig:ecrafter_invSoSy}
\end{figure}
\Cref{fig:ecrafter_invSoSy} shows the driven path of the sensor vehicle exemplarily for one recording.
The duration of the seven single-object recordings averages between \SI{10}{\minute} to \SI{15}{\minute}.

\subsubsection{Multi-object}
The multi-object scenarios address the simultaneous interaction of all seven target vehicles.
\Cref{fig:multiobjekt_pfad} illustrates the test track where they took place. The scenarios can be grouped by the following categories (next to a few independent recordings):

\paragraph{High-dynamic Maneuvers with Frequent Occlusion}
These rather artificial driving maneuvers challenge perception systems with high turn rates and frequent temporary occlusions.
In example, the vehicles drive in figure eights in front of the sensor vehicle, or drive in two circles of different radii around the sensor vehicle.
This poses a particular challenge for (multi-) object tracking algorithms, for example.
Six different scenarios with a total duration of \SI{17}{\minute} were recorded.

\paragraph{Intersection Traffic}
This category of scenarios covers typical intersection traffic.
In this setting, the sensor vehicle stands in front of a crossing and observes the target vehicles crossing in front of it.
The maneuvers are timed such that two-way traffic occurs and causes occlusion.
Two different crossings were selected: one with a good view to the sides of the crossing, and one where the view to the side is occluded by static obstacles.
In the latter case, the target vehicles appear suddenly (and demand quicker track initialization).
This category involves four recordings with a total duration of \SI{5}{\minute}.

\paragraph{Highway and Country Road Traffic}
These scenarios treat typical road and highway traffic. They differ in combinations of the occurrence of approaching traffic, passing maneuvers of the target vehicles and/or the ego vehicle and lane changes. This category contains 13 different scenarios with a total duration of over \SI{28}{\minute}.


\subsection{Provided Data}
All available sensor data is recorded in ROS \emph{bags}. The LiDAR, radar, odometry, and (own and all received target) GNSS/INS sensor measurements are stored as raw data, i.\,e., UDP and CAN packets to reduce the file size and offer the opportunity to process the data with individual data parsers or decoders.
The online documentation gives more information in detail, and offers a ROS \emph{launch file} with suitable sensor data parsers and visualization.
The camera images are stored in a standard ROS format.
Besides, the GNSS/INS data is additionally stored in a processed standard format for simplified data inspection.
The sensor calibration data is supplied by \emph{TF messages} and \emph{CameraInfo messages}, which are also stored in the ROS bags.

In addition to these ROS bags, the raw data dumps of all GNSS/INS units for the complete multi-object measurement campaign are provided.
Finally, the 3D models are also supplied as separate files. The definition of their coordinate systems in relation to the respective GNSS/INS unit is given for each model.

\section{Technical Details and Setup} 
\label{cha:setup}
\subsection{Installation of GNSS/INS units}
\begin{figure}[b]
    \centering
    \includegraphics[width=1\linewidth]{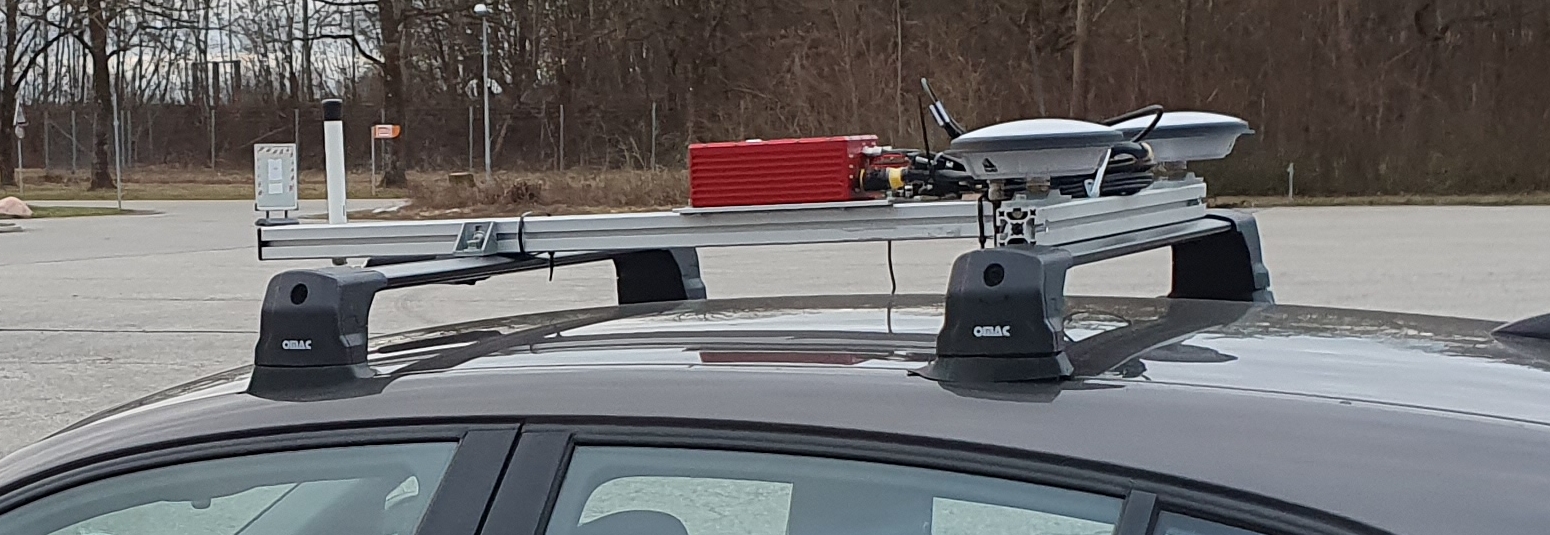}
    \caption{The roof rack carrying the GNSS/INS unit. These roof racks are utilized on the \emph{Hyundai i10}, the \emph{BMW 1 series}, and the \emph{Audi A6}.}
    \label{fig:roof_rack}
\end{figure}
While the prototype vehicles were already equipped with GNSS/INS units, the standard passenger vehicles had lacked a positional reference system.
Thus, we constructed aluminium-based roof racks that host the INS unit, two GNSS antennas, a radio modem to receive differential correction signals from a local RTK-GNSS reference base station, a WiFi device for data communication (sometimes also replacing the radio modem), and power electronics (see \Cref{fig:roof_rack}).
The design focuses on reducing the impact of the construction on the sensor measurements, which, of course, cannot be fully avoided.
It should be noted that the 3D scans were recorded with mounted roof racks, so that the models match the observed measurements.

\subsection{Data Communication}
The sensor vehicle needs to receive the GNSS/INS data of all target vehicles to store and monitor it.
For this purpose, all vehicles were equipped with narrow-band long-range WiFi devices.
The sensor vehicle serves as an access point to all targets.
In most cases, the WiFi range exceeds the relevant sensor range.
However, as this cannot be guaranteed, all GNSS/INS units additionally dump their raw data and solution to an internal storage.
These dumps are also available in this dataset.
In addition, the WiFi connection was also used to relay the DGPS reference data, received by the sensor vehicle, to the targets (in parallel to the DGPS radio).

\subsection{Scanning and Processing of the 3D Models}
\begin{figure}[b]
    \centering
    \includegraphics[width=1\linewidth]{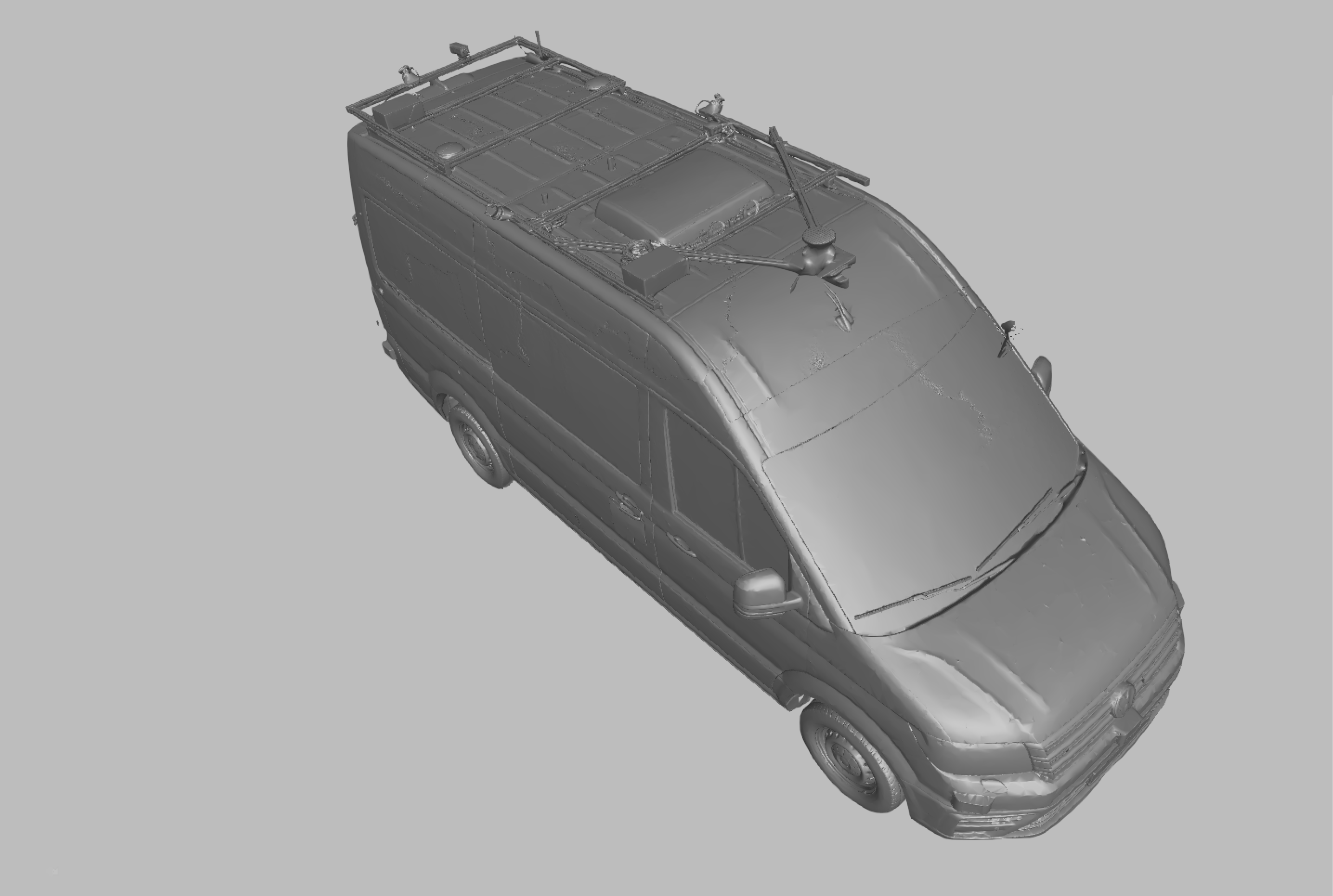}
    \caption{Top view of the scanned 3D model of the \emph{VW e-Crafter} (without texture to be able to distinguish between 3D information and texture).}
    \label{fig:ecrafter_blender}
\end{figure}

All vehicles were scanned with the 3D scanner \emph{Artec Leo} \cite{TAS_Mey2025_PA}.
To improve the scan quality, a texture spray had to be applied on large homogeneous surfaces beforehand.
As this impairs the texture of the model to some extent, its usage was reduced to a feasible minimum.
After postprocessing, a globally registered 3D model is available.
An impression of the quality of the 3D information can be gained in \cref{fig:ecrafter_blender}.
However, at this point, the alignment of its coordinate system is random.
\begin{figure}
    \centering
    \includegraphics[width=\linewidth]{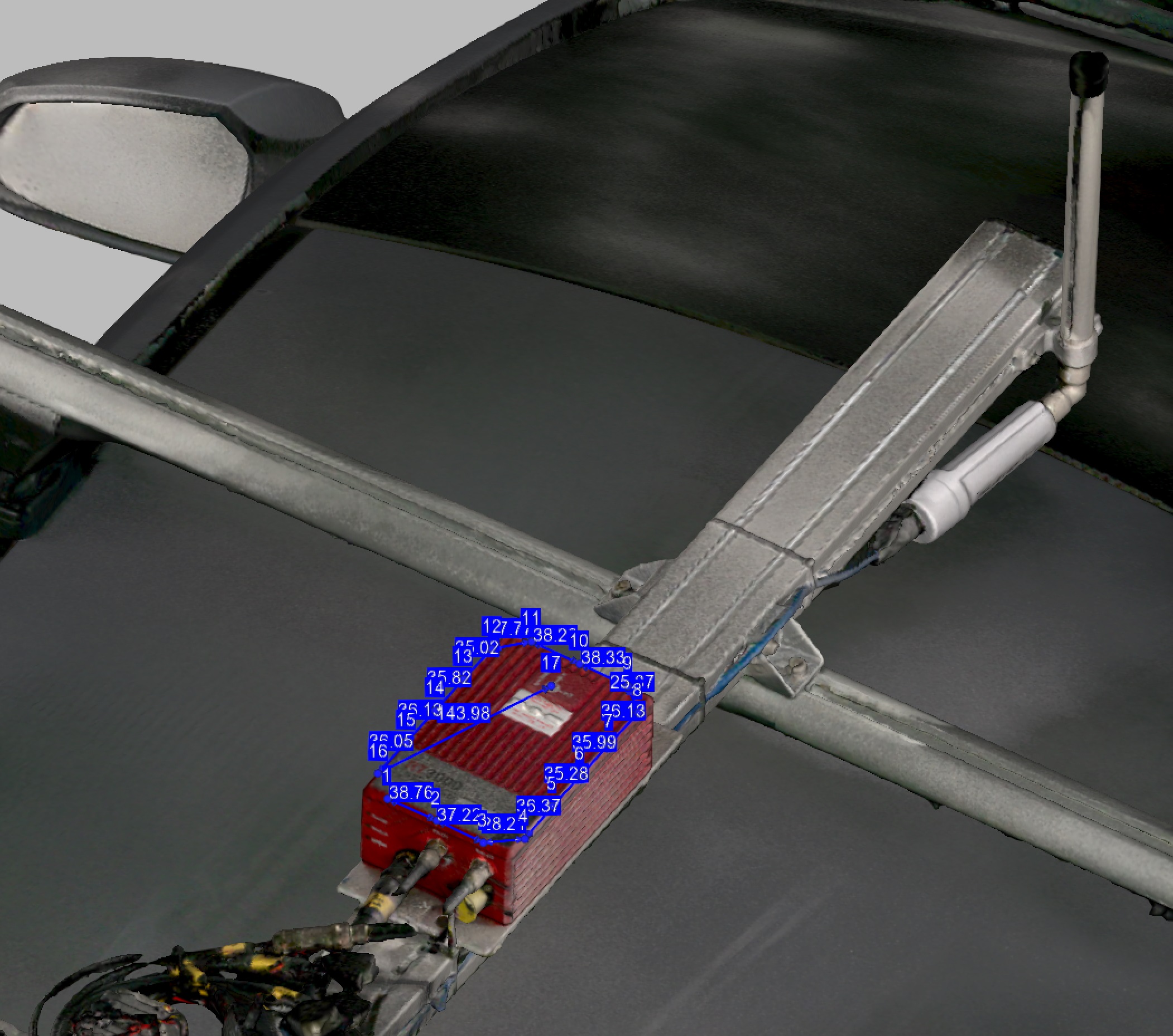}
    \caption{Selection of features in the 3D scan to align the model coordinate system to the GNSS/INS coordinate system. This screenshot shows the selection of screws of an \emph{OxTS RT3000v3} on a roof rack. In this example, the parallelism of the alignment of sets of screws to the axes of the coordinate system has been exploited.}
    \label{fig:oxts_scan}
\end{figure}
To align the coordinate system with the one of the local GNSS/INS unit, some relevant features of the GNSS/INS unit (like markers and screws as shown in \Cref{fig:oxts_scan} whose positions are denoted in its datasheet) or of the prototype vehicle (like axles or emblems) are selected and their positions are extracted from the measured 3D model.
The transformation ${\sideset{^\mathrm{INS}}{}{\mathop \mathbf H}}_{\mathrm{scan}}$, consisting of a translation given by $[x, y, z]$ and a rotation given by $[\Phi, \theta, \Psi]$, converts any point of the measured 3D model $p$ to its equivalent $p'$ in the aligned coordinate system (using rotation matrices $\mathbf R(\cdot)$):
\begin{equation}
    p' = \mathbf{R}_\mathrm z(\Psi) \cdot \mathbf{R}_\mathrm y(\theta) \cdot \mathbf{R}_\mathrm x(\Phi) \cdot  \left(p - (x, y, z)^\intercal \right).
\end{equation}
The converted positions of the selected features can now be evaluated according to known axes coordinates or even positions, symmetry assumptions, or other known alignments. These evaluations are done by cost functions computing a set of $N$ residuals $r_n$. For example, the reference point marker outlines a 3D information (i.\,e., $x$ = \SI{0}{\milli\meter}, $y$ = \SI{0}{\milli\meter}, $z$ = \SI{50}{\milli\meter}) in the INS coordinate system. This point results in three residuals (3D difference between estimated and defined coordinate). A line of $n$ screws arranged in parallel to the INS' x-axis provides $(n-1) \cdot 2$ residuals, i.\,e., the differences between the $y$-coordinates and the differences between the $z$-coordinates should be zero. The optimization problem defined by 
\begin{equation}
    \underset{(x,y,z,\Phi,\theta,\Psi)}{\mathop{\mathrm{arg\,min}}} \sum_{n=1}^N r_n^2
\end{equation}
finally yields the desired transformation.

\subsection{Accuracy Estimates of the Transformation Chain} 
\label{sec:accuracy}
The overall projection error, that occurs when projecting the 3D model into a sensor's coordinate system, depends on multiple factors. To break this down, we regard the transformation chain ${\sideset{^\mathrm{sensor}}{}{\mathop \mathbf H}}_{\mathrm{scan}}$, that transforms a point $p$ in the 3D scan to a point $p''$ in the sensor coordinate system:
\begin{equation}
    \begin{split}
    p''(t) =& {\sideset{^\mathrm{sensor}}{}{\mathop \mathbf H}}_{\mathrm{scan}} (t) \cdot p \\
           =& {\sideset{^\mathrm{sensor}}{}{\mathop \mathbf H}}_{\mathrm{ego-INS}} \cdot
              {\sideset{^\mathrm{ego-INS}}{}{\mathop \mathbf H}}_{\mathrm{target-INS}} (t) \cdot \\
            & {\sideset{^\mathrm{target-INS}}{}{\mathop \mathbf H}}_{\mathrm{scan}} \cdot p.
    \end{split}
\end{equation}
The sensor-to-INS calibration ${\sideset{^\mathrm{sensor}}{}{\mathop \mathbf H}}_{\mathrm{ego-INS}}$ consists of a precise data-driven sensor calibration and the determination of the GNSS/INS position using CAD documents.  
The accuracy of the relative vehicle pose ${\sideset{^\mathrm{ego-INS}}{}{\mathop \mathbf H}}_{\mathrm{target-INS}}$ provided by the GNSS/INS units is indicated as a covariance matrix in the data stream. The manufacturer states a positional accuracy of \SI{1}{\centi\meter} and an angular accuracy of \SI{0.03}{\degree} for each unit under perfect conditions (to obtain the accuracy for a relative pose between two units, these parameters need to be applied twice). The recordings were settled under open sky conditions with a nearby RTK-GNSS reference station, resembling very good conditions. In addition, long RTK-warm-up maneuvers were performed before the measurement campaign. A dedicated check finally confirmed the given accuracies.
The accuracy of the model alignment ${\sideset{^\mathrm{target-INS}}{}{\mathop \mathbf H}}_{\mathrm{scan}}$ is strongly coupled with the accuracy of the scan itself, i.\,e., the trueness of the selected point $p$ of the scan. The accuracy of the scan is -- according to the manufacturer -- $\SI{0.1}{\milli\meter}$ plus $\SI{0.3}{\milli\meter}$ per meter of scan distance. As this accuracy can actually be obtained for ``simple'' objects (which ironically correspond to objects with varying structure and texture for simplified scan registration), the correct registration can be tough for homogeneous surfaces. As long as the registration has been performed correctly, the obtained accuracy is indeed at least in the range of few centimeters according to a comparison with other measurement approaches \cite{bifo:TAS_Deuter2022_PA}.

\subsection{Auto-Annotation}
We do not provide ready-to-use annotations as they strongly depend on the desired annotation format and the individual application. To automatically create annotations with custom format and granularity, ray casting libraries like \cite{Dawson2019trimesh} may be used.

\section{Measurement Evaluation Examples}
\label{cha:examples}
This section briefly presents some exemplary data evaluation results.

\begin{figure}[b]
    \centering
    \includegraphics[width=0.6\linewidth]{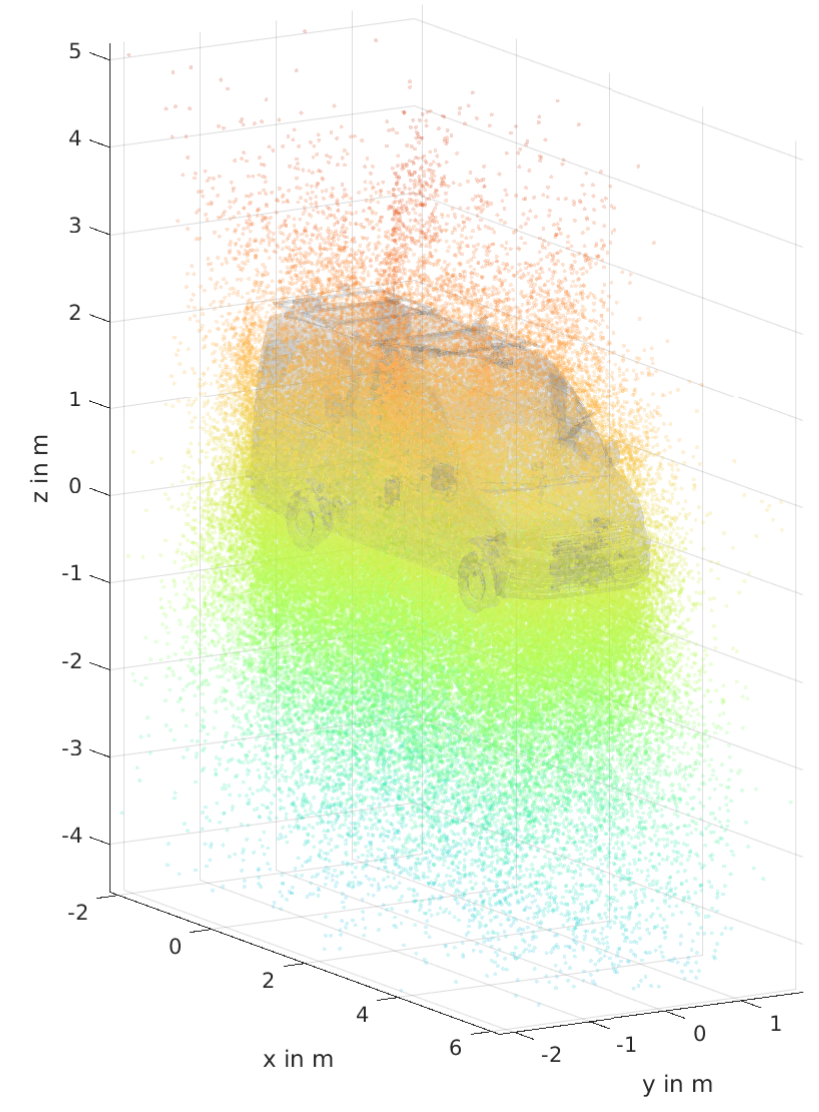}
    \caption{Accumulation of radar detections in the coordinate system of the target. This rendering shows such an accumulation of the \emph{VW e-Crafter} together with the aligned alpha-blended 3D model.}
    \label{fig:ecrafter_detektions_akkumulation}
\end{figure}
To evaluate the spatial occurrence of measurements, an accumulation of point detections can be performed in the coordinate system of a target. \Cref{fig:ecrafter_detektions_akkumulation} shows such an accumulation using the radar detections.
In our case, it manifests a large uncertainty in the elevation measurements of our radar sensors, for example.

A more quantitative evaluation is the computation of heatmaps. The approach shown in the following registers the occurrences of radar detections in a 2D (i.\,e., top-down) heatmap. Every time a sensor performs a measurement, the list of detections is converted to a target-fixed grid. Besides, the respective observation parameters (distance and angle) are attached to the grid as meta information. The accumulation of these grids results in a heatmap. However, these heatmaps would be subject to an observation bias: if a specific relative pose has been recorded more often than another one, it would also have a higher influence on the heatmap. This effect is undesired as it would wrongly imply a higher static reflectivity of the observed part of the object. Therefore, we utilize a normalization algorithm \cite{Berthold2017} that weights these grids in inverse relation to the frequency of the occurrence of their observation parameters.
\begin{figure}
    \centering
    \begin{subfigure}{1\linewidth}
        \centering
        \includegraphics[width=\linewidth]{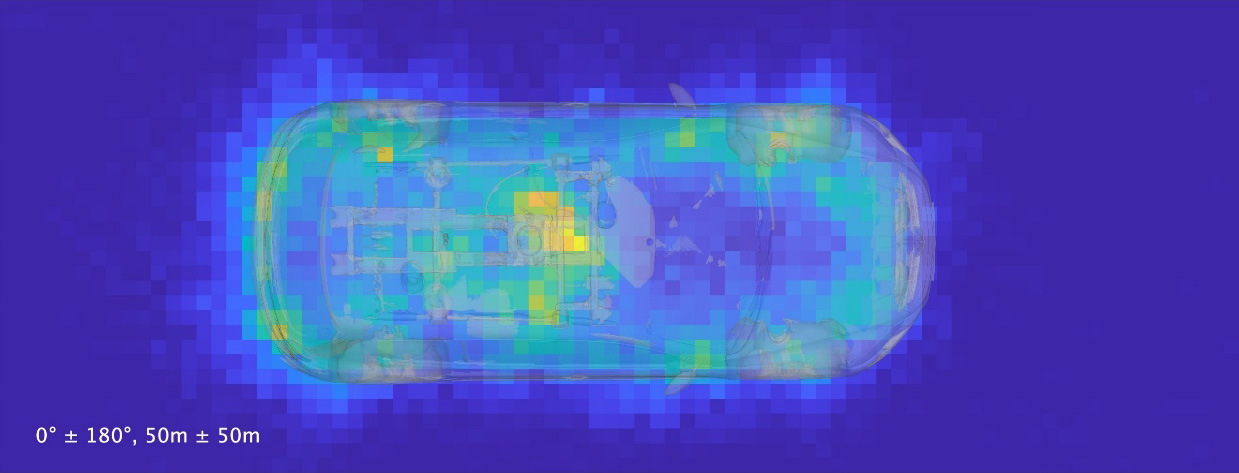}
        \caption{Heatmap rendered over all observations.}
        \label{fig:heatmap:full}
    \end{subfigure}
    \begin{subfigure}{1\linewidth}
        \centering
        \includegraphics[width=\linewidth]{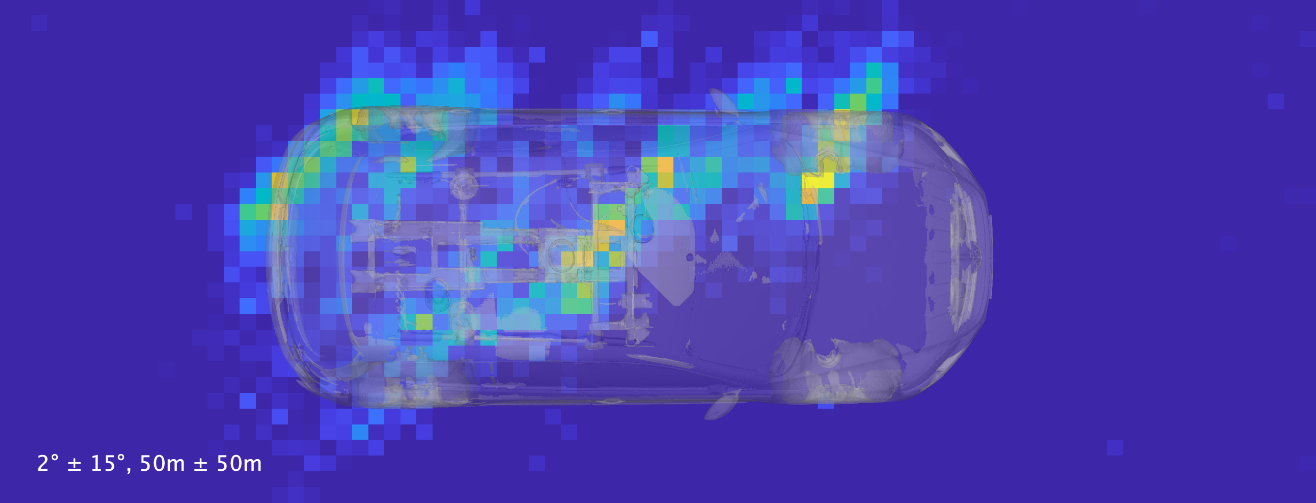}
        \caption{Heatmap rendered for all measurements observed when the sensor has been located in the rear left of the target vehicle. It can be easily spotted that most detections originate from the rear left corner, the roof structure (\emph{Velodyne VLS-128}), and both left wheels. In addition, the angular measurement noise is apparent.}
        \label{fig:heatmap:linksOben}    
    \end{subfigure}%
    \caption{Renderings of normalized radar detection heatmaps with overlays of the scanned 3D model for inspection purposes. These heatmaps are rendered on base of the \emph{VW Tiguan} recording. Dark blue cells indicate a low occurrence of detections, yellow cells indicate a high occurrence.}
    \label{fig:heatmap}
\end{figure}
The overlay of the 3D model allows for the association of specific hotspots to their real counterparts. \Cref{fig:heatmap:full} shows such an illustration. It reveals hotspots along the contour of the vehicle and on the sensor setup on the roof. It is also possible to select specific observation parameters (e.\,g. all measurements from the front of the target) to analyze view-dependent heatmaps as shown in \Cref{fig:heatmap:linksOben}.

\begin{figure}
    \centering
    \includegraphics[width=1\linewidth]{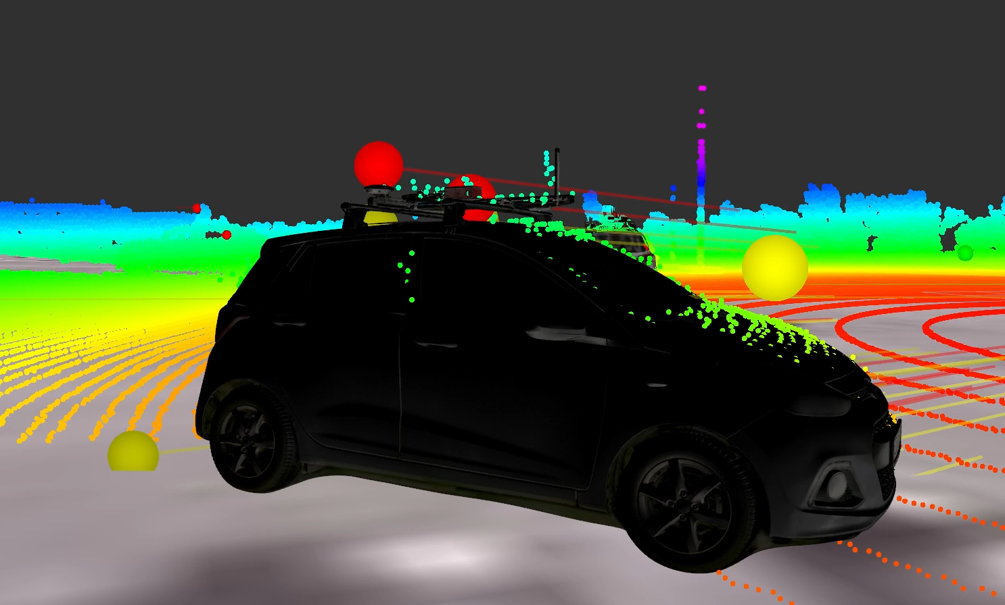}
    \caption{LiDAR points measured at the opposite b-pillar from the sensor. The laser ray permeated the window, obviously. The LiDAR point cloud is illustrated by the small spheres; their color-coding depends on the range. The yellow/red spheres again denote radar detections.}
    \label{fig:i10_b-saeule}
\end{figure}

Similar evaluations can also be performed for the LiDAR sensor. An interesting outcome is shown in \Cref{fig:i10_b-saeule}.
With the help of the 3D model, the permeability of car windows for LiDAR points can be examined.
\begin{figure}
    \centering
    \includegraphics[width=1\linewidth]{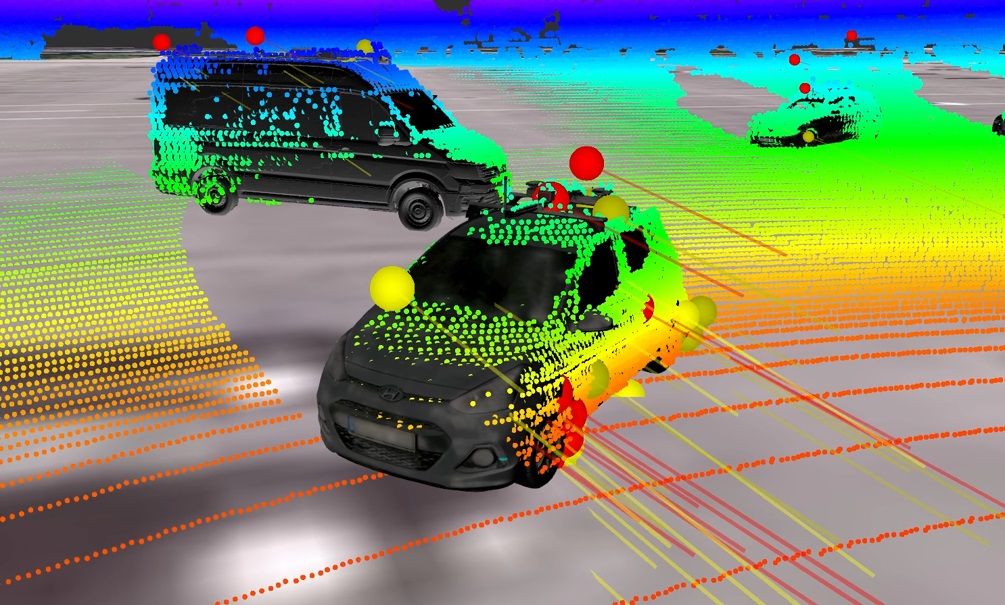}
    \caption{The \emph{Hyundai i10} partly occludes the \emph{VW e-Crafter} in the background. The lines of the red/yellow spheres denote the radial speed / Doppler measurement and are radial to the respective radar sensor.}
    \label{fig:occlusion}
\end{figure}
Finally, \Cref{fig:occlusion} shows the effect of occlusion on the measurements. By knowing the exact 3D model and thus the missing sensor data, the behavior of algorithms in occlusion can be precisely investigated and enhanced.

\section{Conclusion, Use Cases, and Future Work}
\label{cha:conclusion}
This paper presents a multi-vehicle dataset with accurate 3D models of all vehicles involved to improve the modeling or learning of physical and technical measurement principles and the appearance of vehicles in measurement data. To this end, we have scanned all vehicles using a 3D scanner, equipped all vehicles with GNSS/INS units, and aligned these models. The contributed concept allows for the auto-annotation of all sensor measurements with custom granularity, as the state, the shape, and the structure of all objects are known for any point in time. This helps to develop, improve, train, or evaluate perception algorithms based on camera images, LiDAR point clouds or radar detections with individual needs for detail. As we utilize seven known objects, even advanced effects like occlusion and multi-path reflections are covered by the ground truth information.
Ultimately, the 3D models can also be used on their own to simulate any scenario with realistic vehicle models.

In addition, the dataset provides dense and accurate kinematic ground truth information for all vehicles involved.
Thus, it can also be used to evaluate (multi) object tracking algorithms.
In longer scenarios, even odometry or SLAM algorithms could be evaluated.

To obtain positional and kinematic reference data, this approach utilizes the installation of GNSS/INS units on roof racks. Although the scanned 3D models incorporate these structures and therefore match the measurements, it might be beneficial for subsequent datasets to use hidden GNSS antennas and internal IMUs (or other localization techniques). In addition, integrating joints in the models to allow for moving wheels might be reasonable future work.

\section*{Acknowledgement}
This work is funded by the Federal Office of Bundeswehr Equipment, Information Technology and In-Service Support (BAAINBw).
The authors would like to thank Moritz Mey for performing the 3D scans and their post-processing, and Anton Backhaus, Alexander Bienemann, Thorsten Luettel, Martin Maier, Arian Rottmann, Thomas Rottmann, Thomas Steinecker, and Georg Thalhammer for participating in the recording of the multi-vehicle scenarios. The authors thank Thorsten Luettel additionally for conducting the scenarios. Finally, the authors thank their private cars for taking some scratches for the sake of science.

\bibliographystyle{IEEEtran}
\bibliography{macros/IEEEabrv,macros/additional_abrv,bibliography}

\begin{acronym}
\acro{SRR}{Short-Range Radar}
\acro{SPRB}[SPRB]{Set of Points on a Rigid Body}
\acro{RFS}{Random Finite Set}
\acro{COTS}{Commercial Off-The-Shelf}
\acro{CFAR}{Constant False Alarm Rate}
\acro{CUT}{Cell Under Test}
\acro{CTRV}{Constant Turn Rate and Velocity}
\acro{SNR}{Signal-to-Noise ratio}
\acro{PDAF}{Probabilistic Data Association Filter}
\acro{JPDAF}{Joint Probabilistic Data Association Filter}
\acro{RMSE}{root-mean-square error}
\acro{RCS}{Radar Cross Section}
\acro{radar}{radio detection and ranging}
\acro{ACC}{Adaptive Cruise Control}
\acro{MAP}{maximum a posteriori}
\acro{MHT}{Multiple Hypothesis Tracking}
\acro{VGM}{Variational Gaussian Mixtures}
\acro{LMB}{Labeled Multi-Bernoulli Filter}
\acro{PMHT}{Probabilistic Multi-Hypothesis Tracking}
\acro{SUV}{Sport Utility Vehicle}
\acro{MSE}{mean square error}
\acro{ADAS}{Advanced Driver Assistance Systems}
\acro{EKF}{Extended Kalman Filter}
\acro{RMSE}{Root-Mean-Square Errors}
\end{acronym}

\end{document}